\documentclass{article}

\usepackage{microtype}
\usepackage{graphicx}
\usepackage{subfigure}
\usepackage{amsmath}
\DeclareMathOperator*{\argmax}{arg\,max}
\usepackage{multirow}
\usepackage{booktabs} % for professional tables

\usepackage{hyperref}

\usepackage[accepted]{icml2021}

\icmltitlerunning{Lagrangian Objective Function Leads to Improved Unforeseen Attack Generalization in Adversarial Robustness}

\begin{document}

\twocolumn[
\icmltitle{Lagrangian Objective Function Leads to Improved Unforeseen Attack Generalization in Adversarial Training}



\icmlsetsymbol{equal}{*}

\begin{icmlauthorlist}
\icmlauthor{Mohammad Azizmalayeri}{to}
\icmlauthor{Mohammad Hossein Rohban}{to}
\end{icmlauthorlist}

\icmlaffiliation{to}{Department of Computer Engineering, Sharif University of Technology, Tehran, Iran}

\icmlcorrespondingauthor{Mohammad Azizmalayeri}{azizmalayeri@ce.sharif.edu}
\icmlcorrespondingauthor{Mohammad Hossein Rohban}{rohban@sharif.edu}

\icmlkeywords{Deep Learning, Robustness, Attack, Lagrange, Defence}

\vskip 0.3in
]



\printAffiliationsAndNotice{}  % leave blank if no need to mention equal contribution

\begin{abstract}
Recent improvements in deep learning models and their practical applications have raised concerns about the robustness of these models against adversarial examples. 
%Since our knowledge about possible adversarial attacks is limited and new powerful attacks may be proposed at any moment, we should use training algorithms that lead to intrinsic robustness of models and makes models robust against unforeseen attacks. 
Adversarial training (AT) has been shown effective to reach a robust model against the attack that is used during training. However, it usually fails against other attacks, i.e. the model overfits to the training attack scheme.  %due to restricting training attack by avoidable constraints so the model can't learn possible features of other attacks during training. 
%Here we develop a new adversarial attack without useless constraints that is specifically designed to be used in adversarial training to generate a robust model. 
In this paper, we propose a simple modification to the AT that mitigates the mentioned issue. More specifically, we minimize the  perturbation $\ell_p$ norm while maximizing the classification loss in the Lagrangian form. We argue that crafting adversarial examples based on this scheme results in enhanced attack generalization in the learned model.
We compare our final model robust accuracy against attacks that were not used during training to  closely related state-of-the-art AT methods. This comparison demonstrates that our average robust accuracy against unseen attacks is 5.9\% higher in the CIFAR-10 dataset and is 3.2\% higher in the ImageNet-100 dataset than corresponding state-of-the-art methods. We also demonstrate that our attack is faster than other attack schemes that are designed for unseen attack generalization, and conclude that it is feasible for large-scale datasets\footnote{Code is available at \url{https://github.com/rohban-lab/Lagrangian_Unseen}.}.
\end{abstract}

\section{Introduction}\label{Introduction}

Deep neural networks have been used in many applications and have achieved impressive results in various domains, especially in computer vision tasks such as image classification, autonomous vehicles, and face recognition \cite{Lin2018, Wang2020}. But there are still a lot of security-based concerns about them as it has been demonstrated that these networks are not sufficiently robust against different types of adversarial examples \cite{Akhtar2018, Szegedy2014}. Adversarial examples are data points that are slightly perturbed and optimized to deceive a network. Numerous defenses have been developed against adversarial examples, but the issue has not yet been completely solved as the proposed methods lead to robustness against a limited number of threat models with specific conditions.\newline
Variants of adversarial training (AT) \cite{Madry2018} are currently state-of-the-art algorithms among empirical defenses. AT tries to solve an optimization problem of minimizing loss of the network \emph{f} parameterized by \emph{\(\theta\)} based on adversarial examples \emph{\(x_i + \delta_i\)}. The perturbation $\delta_i$ is obtained by maximizing the network loss for $x_i+\delta_i$. In order to make the perturbation imperceptible, \emph{\(\delta_i\)} could be norm bounded. Therefore, AT briefly involves:
\[\min_\theta \sum_i \max_{\delta \in \Delta} \ell(f_\theta(x_i + \delta), y_i),\]
where $\ell(.)$ is the loss function, and $\Delta$ is the set of feasible perturbations.
The outer minimization can be done using the stochastic gradient descent algorithm, but the challenging part often involves solving the inner maximization. %In other words, we have a min-max game between \emph{\(\delta\)} that is optimized to maximize network loss and \emph{\(\theta\)} that is being optimized to minimize loss to defeat attacks similar to \emph{\(\delta\)} in the game. 
It has been empirically observed that the  final model will be robust mainly against perturbations within $\Delta$, but almost fails under other threat  models \cite{Ilyas2019}.

To tackle this challenge, two strategies were considered in the literature: training against union of perturbation sets \cite{Miani2020}, or using a more perceptually aligned metric to define the perturbation set, $\Delta$ \cite{Laidlaw2020}. Along the first strategy, \cite{Miani2020} proposed an attack that is updated based on the gradient from the worst case error of \emph{\(\{\ell_1, \ell_2, \ell_\infty\}\)} attacks in each iteration. Such methods do not still generalize to imperceptible attacks that are not norm bounded \cite{Laidlaw2020}. 
As an alternative, it was proposed to broaden $\Delta$ to contain more general attacks. For instance, \cite{Laidlaw2020} recently used a deep neural network to bound the attack based on the network embedding, which is called the LPIPS distance and is believed to be a more perceptual metric. The constrained version of their algorithm is computationally demanding, so they proposed a Lagrangian penalty in designing the attack. This idea led to promising results on unseen attacks. 

We will build upon this work and show that the Lagrangian nature of the loss plays a more important role in unseen attack generalization than the LPIPS distance. We observed that the results improve further if we replace the LPIPS in the Lagrangian setting with the $\ell_2$ distance! We give some theoretical insights on a possible explanation for this observation. It has also been pointed out that due to the deep neural network adversarial fragility, the LPIPS metric has certain shortcomings in modeling the perceptual distances \cite{Kettunen2019}.

\begin{figure}[t]
\vskip 0.05in
\begin{center}
\centerline{\includegraphics[width=\columnwidth]{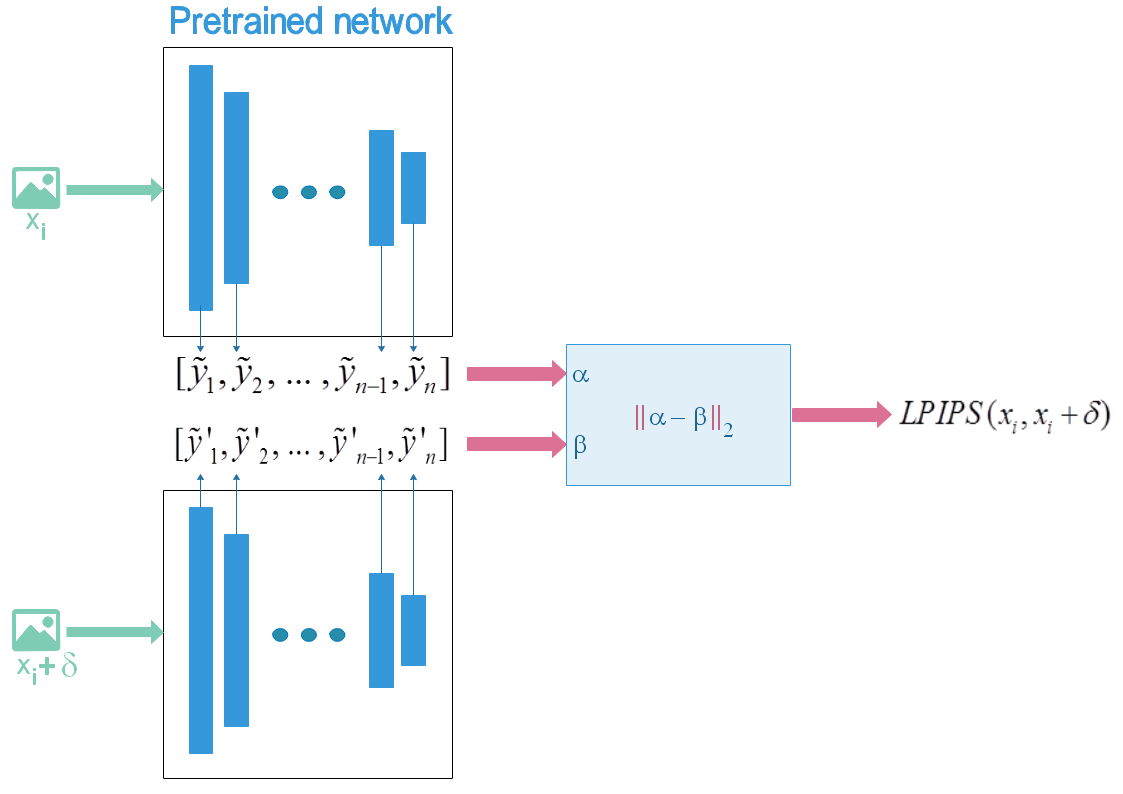}}
\vskip 0.05in
\caption{LPIPS distance. First, outputs of convolutional layers of a pretrained network for each image is calculated \emph{\( (y_i, y_{i}^{\prime} ) \)} and normalized using height and width \emph{(\(\tilde{y}_i, \tilde{y}_i^{\prime}\))}. Next, $\ell_2$ norm difference between \emph{\(\tilde{y}\)} and \emph{\(\tilde{y}^{\prime}\)} is calculated that gives LPIPS distance between images.}
\label{LPIPS distance}
\end{center}
\vskip -0.2in
\end{figure}

%One of the widely used constraints that is used in AT is bounding the attack \emph{\(\ell_p\)} norms \cite{Miani2020, Croce2019, Madry2018}. The caveat is that an imperceptible adversarial attack may have a large \emph{\(\ell_p\)} norm. \cite{Laidlaw2020} considered this fact and used LPIPS distance \cite{zhang2018} as a perceptual distance metric in crafting the attack.  
%If we completely remove projection, we will suffer from perturbations with large amplitudes. Our solution to this problem is telling gradients that they can not be too large! In other words, we add a penalty to the loss function for calculating excessive large \emph{\(\delta\)}. So our adversarial training would become:
%\[\min_\theta \sum_i \max_\delta (loss(f_\theta(x_i + \delta), y_i) - \lambda \ distance(x_i , x_i + \delta))\]
%\cite{Laidlaw2020} used similar method to propose Fast-LPA attack but instead of \emph{\(distance(x_i , x_i + \delta)\)}, they used \emph{\(Relu(distance(x_i , x_i + \delta) - \gamma)\)} that may be helpful in test time but it is forcing \emph{\(\delta\)} to reach \emph{\(distance(x_i, x_i+\delta) \approx \gamma\)} which is a restriction for \emph{\(\delta\)} and prevents the attack generality in training time.\newline
We also note that in the ideal case, the distance metric should not have any a priori bias towards specific \emph{\(\delta\)}. This fact brings out an issue with the LPIPS distance. As demonstrated in Fig. \ref{LPIPS distance}, LPIPS uses the $\ell_2$ norm difference of outputs of internal convolutional layers of a pre-trained network for calculating the distance between \emph{\(x_i\)} and \emph{\(x_i+\delta\)}. Internal layers of the networks are mainly sensitive to foreground, i.e. parts of the image that contain the target object. As a result, changes in the background of the object do not often change the internal layers a lot. This would encourage the perturbation to make changes in the background and cause some natural bias in the attack. We will discuss this bias in more details in section \ref{Experiment}.

According to these reasons, we suggest $\left\| \delta \right\|_2$ as the distance metric since it does not impose such biases in the attack. Closely related to this attack, C\&W attack \cite{Carlini2017} also tries to find the minimal perturbation that can fool the classifier by adding a second term to the loss function to avoid large perturbations. %But instead of directly optimizing $\left\| \delta  \right\|_p$ like what we do in our work, they defined \emph{\(\delta=\dfrac{1}{2}(\tanh(\omega)+1)-x\)} and optimized \emph{\(\omega\)} which is not really necessary for our goal. 
However, the main difference of \cite{Carlini2017} with our work is in adjusting the weight, \emph{\(\lambda\)}, that is multiplied by the norm in the loss. For each instance, C\&W attack initializes \emph{\(\lambda\)} with a large value and gradually decreases it until it reaches a successful adversarial example. This gives us the minimal successful perturbation, which sounds reasonable for the test time. But it has two disadvantage in the training time. First, searching over \emph{\(\lambda\)} makes this attack infeasible for large-scale datasets. The second disadvantage is the fact that other attacks do not necessarily generate minimal perturbations, and hence C\&W does not make the model robust against such attacks. Therefore, we propose to use a fixed schedule of \emph{\(\lambda\)} for all samples to avoid these problems. More importantly, we argue that based on the envelope theorem, a {\it fixed} $\lambda$ places an upper bound on the gradient of the loss with respect to the perturbation budget, $\epsilon$, while also prevents  $\delta$ from be overly small on average. Therefore, we hypothesize that fixing $\lambda$ prevents the loss function to change drastically across slightly different perturbation sets, and results in improved unforeseen attack generalization. 

%Another widely used constraint is using gradient sign \cite{Kurakin2017, Goodfellow2015}. The main reason for using gradient sign is preventing gradients to have large amplitudes but we already have solved the problem of large gradient amplitude so we can use gradient itself instead of the sign of gradient. By using gradient itself, in addition to removing an avoidable constraint, we can reach the optimal \emph{\(\delta\)} faster since we are taking steps in proportion to gradient amplitude and our steps can be short or long according to the gradient in each point. With this benefit, we could reduce the essential steps for optimizing \emph{\(\delta\)} so our proposed attack and our adversarial training would be much faster than similar methods which would be discussed in section \ref{Experiment}.
In summary, we propose to use Lagrangian objective function with a fixed multiplier to achieve better average accuracy against unforeseen attacks. We give some theoretical insights to support the use of such attacks in AT. We also note that the mentioned attack is feasible for large datasets. We finally compare our method with state-of-the-art methods against attacks that were not used during training, and will demonstrate that our method is outperforming state-of-the-art.

\section{Related Work}

Adversarial attacks and defenses have attracted a lot of attention in recent years. Attacks are being proposed sequentially while defenses try to confront them and this loop always exists \cite{Uesato2018, Athalye2018}.
%Here we first explore popular adversarial attacks. After that, empirical defenses are investigated.
Here, we explore some popular adversarial attacks and empirical defenses.
\subsection{Adversarial attacks}
One of the first successful attacks against deep neural networks was the Fast Gradient Sign Method \cite{Goodfellow2015} that used input gradient signs to craft the adversarial perturbation. \cite{Madry2018} tried to perform FGSM iteratively inspired by \cite{Kurakin2017} and proposed Projected Gradient Descent (PGD) that moves iteratively along the gradient direction, and scale the perturbation using the \emph{\(\ell_p\)} norms.  MI-FGSM \cite{Dong2018}  used momentum to modify the gradient direction in each step of designing the perturbation. %\cite{Papernot2016} used jacobian saliency map
%as a metric to measure more effective points of input in output and used more important points 
%to reduce model accuracy.
%\cite{Carlini2017} proposed CW attack and achieved suitable results against defensive distillation but their attack is not feasible for training time.
\cite{Rony2020} used the Lagrange method, similar to the C\&W attack, \cite{Carlini2017} to generate minimal perturbations, except that they modify the Lagrange weight in each iteration in a way that their distance metric is better satisfied.
DeepFool attack assumed that the classifier is linear in a neighborhood of each example and tried to
reduce the model's accuracy by moving the example toward the nearest linear border \cite{Moosavi2016}. \emph{\(\ell_0\)} attacks such as PGD$_0$ and CornerSearch, deceive network with changing minimal pixels \cite{Croce2019, Apostolos2019}.
%attack model with minimum \emph{\(L_2\)} norm of perturbation \cite{Moosavi2016}.
%Mentioned attacks try to find perturbation for a specific sample but \cite{Moosavi2017} proposed Universal adversarial attack that aims to craft a single perturbation to fool model on most of the samples. 

As an alternative to the $\ell_p$ norm bounded attacks, Wasserstein distance for adversarial robustness was introduced by \cite{Wong2020} as a better distance for imperceptible perturbations. However, their proposed Sinkhorn iterations for projecting onto the Wasserstein ball is computationally expensive and prohibitive in large datasets. Perceptual metrics such as SSIM \cite{Wang2004} and LPIPS \cite{zhang2018} are another alternatives for the standard \emph{\(\ell_p\)} norms. For instance, \cite{Laidlaw2020} proposed the LPA and PPGD attacks using the LPIPS distance, Fig. \ref{LPIPS distance}, for crafting perturbations that are not visible and recognizable for human. 

As the perturbation set is broader in such attacks, compared to the traditional $\ell_p$ attacks, one would expect robustness against unseen attacks once the model is trained using those attacks. In order to check the model's accuracy against unseen attacks, \cite{Kang2020} proposed JPEG, Fog, Snow, and Gabor attacks that we will also use to compare our final model's accuracy with the rest.
%Another different type of adversarial attack is patch attack that perturbs a segment of the image instead of pixels, for example, \cite{Sharif2016} added eyeglass to human faces to fool the classifier. 
%Generative models are also used for creating adversarial examples via learning the adversarial distribution but they still have instabilities problems \cite{Song2018}.
 StAdv \cite{xiao2018} and RecolorAdv \cite{Laidlaw2019} are also recently proposed as other alternative adversarial attacks. StAdv makes local spatial transformation of each input pixel adversarially, and RecolorAdv maps the pixel original color to a perceptually indistinguishable color to fool the classifier. We will also evaluate our model based on these attacks.
%In our work, we propose an attack without mentioned constraints such as using the sign of gradient or projecting perturbation via \emph{\(L_p\)} norms, Wasserstein distance, perceptual metrics, etc. or any other specific assumptions which makes our attack suitable for the training time.

\subsection{Empirical Defenses}
%Auto-encoders were proposed as a defense against adversarial examples in order to extract better features from images \cite{Schott2019}. 
Several defense methods have been proposed in the literature, but as pointed out in \cite{Athalye2018b}, many of these defenses suffer from a phenomenon called ``gradient masking" and ``gradient obfuscation." Perhaps the most  effective defenses are variants of adversarial training \cite{Madry2018}. This includes training with PGD adversarial examples. Our work is also concentrated on the same concept. 

Adversarial training on multiple perturbations is also investigated to reach robustness against multiple threat models \cite{Trame2019, Miani2020}, but our work is more general and aims to reach robustness against all unseen attacks.
%\cite{Wang2020Improving} considered misclassified examples and tried to bring output distribution of misclassified examples closer to output distribution of clean examples.
Adversarial training can also be combined with self supervised learning methods to improve robustness \cite{Kim2020, Chen2020}. But in this work, we try to improve the base adversarial training, which could lead to improvement of its variants. Finally, note that there could be a trade-off between standard accuracy and robust accuracy \cite{tsipras2019}. Therefore, to have a fair evaluation, one has to compare the adversarial accuracy of different methods, under the same or similar clean accuracy. 

\section{Proposed approach} \label{proposed}
The main contribution of our work is proposing an attack that is specifically designed to be used in adversarial training \cite{Madry2018}. In our proposed attack, we attempt to maximize \emph{\(\ell(f_\theta(x_i + \delta), y_i) \)}, where \emph{f} is the network parametrized by \emph{\(\theta\)} that classifies samples \emph{\(x_i\)} with labels \emph{\(y_i\)}, and we use \cite{Carlini2017} margin loss as ${\ell(.)}$. In order to force the perturbation \emph{\(\delta\)} to be small, we use the Lagrangian formulation and add a penalty term $\| \delta \|_2$ to the loss function. We will empirically show that this simple modification boosts the generalization of the trained model to new unforeseen threat models. 

As mentioned in section \ref{Introduction}, the penalty term should preferably be unbiased against penalizing specific perturbations at the training time. Otherwise, it will cause overfitting to the training attack and prohibits the generalization of the classifier to unforeseen threat models. To consider this issue, we suggest \emph{\({\| \delta \|}_2\)} as the penalty term, since it treats all pixels of the image in the same way. Other differentiable $\ell_p$ norms can also be used for this purpose, but the optimization may become harder. So our final objective function for generating perturbations would be:
%Considering discussed issues in section \ref{Introduction}, to keep our attack more general, we use \emph{\({\lvert\lvert \delta \rvert\rvert}_2\)} as second term of the loss function. So our attack final loss function would be:
%\vskip 0.1mm
%\[\max{(\max_{i\neq y}(f_\theta (x + \delta)_i - f_\theta (x + \delta)_y), -\kappa)} - \lambda \ \emph{\({\lvert\lvert \delta \rvert\rvert}_2\)}\] %where \emph{\(f_\theta (x)_i\)} is the \emph{\(i_{th}\)} logit of network \emph{\(f_\theta\)} for input \emph{x} and $\kappa$ is a parameter to control the misclassified examples confidence.
\[\max_\delta \ell(f_\theta(x_i + \delta), y_i)-\lambda \ \emph{\({\| \delta \|}_2\)},\]
and $\lambda$ is the constant Lagrange multiplier that determines the compromise between the first and second terms. The penalty term acts as a ``perturbation decay" in gradient descent (GD) iterations, and prevents the input gradients from over-enlarging in each iteration.
%Adding the penalty term in addition to the mentioned benefit controls the amplitude of the gradient so we can move toward pure gradient instead of the sign of gradient.
%We use GD with five steps to perform this optimization.
%However, using the fact that perturbation doesn't go further away from its initialization, we don't use any projections to keep the generality of our attack.

%As we simultaneously minimize \emph{\({\| \delta \|}_2\)}, we avoid overly large $\delta$ perturbations. For this reason, unlike previous work, we do not project our final perturbation to a specific space.
%To accelerate the optimization, \emph{\(\lambda\)} is exponentially increased, and the learning rate is exponentially decayed. In order to make gradients from different samples on the same scale, we normalize the gradient from each sample by the maximum of absolute gradients in different points of that sample.
At least five steps are required to perform this optimization since we maximize the loss and minimize the distance simultaneously. Using five steps, we can start with a smaller $\lambda$ to move a sample in a space with maximum loss without worrying about the distance. In the following steps, $\lambda$ is increased to make the $\ell_2$ distance smaller, and simultaneously reduce $\alpha$ to make sure that more minor changes are made in that step, and the sample is still adversarial. To ensure that decreasing $\alpha$ reduces the amount of changes, the gradient scale in different steps should be the same. So, we normalize the gradients in each step.
Our attack is summarized in Alg. \ref{General attack}. 
\begin{algorithm}[tb]
   \caption{Our adversarial attack designed specifically for adversarial training}
   \label{General attack}
\begin{algorithmic}
   \STATE
   \STATE {\bfseries Input:} network classifier $f_\theta$, data $x$, label $y$
   \STATE {\bfseries Parameters:} number of iterations $N$,
   %margin loss confidence control $\kappa$,
   initial step size $\alpha$, initial Lagrange multiplier  $\lambda$, perturbation initialization variance $\sigma^2$ (default 0.01), decay parameter $c$ (default 0.1)
   \STATE {\bfseries Output:} perturbation $\delta$
   \STATE
   \STATE $\delta \sim \mathcal{N}(0, \sigma^2)$
   %\STATE $\tilde{x} = x + \delta$
   \FOR{$i=0$ {\bfseries to} $N-1$}
   \STATE $\lambda^{\prime} \leftarrow \lambda \times c^{1-i/N}$
   \STATE $\alpha^{\prime} \leftarrow \alpha \times c^{i/N}$
   %\STATE $Margiloss = max(max_{j\neq y}f_\theta (x + \delta)_j - f_\theta (x + \delta)_y, -\kappa)$
   %\STATE $error = \ell(f_\theta(x + \delta), y)$
   %\STATE $error = loss(f_\theta(\tilde{x}), y)$
   %\STATE $Penalty\ loss= \emph{\({\lvert\lvert \delta \rvert\rvert}_2\)}$
   %\STATE $penalty = \emph{\({\lvert\lvert \delta \rvert\rvert}_2\)}$
   %\STATE $gradients = \nabla (Margin\ loss + \lambda^{'} Penalty\ loss)$
   \STATE $g \leftarrow \nabla (\ell(f_\theta(x + \delta), y) - \lambda^{\prime} \| \delta \|_2)$
   \STATE $g \leftarrow g/\max(|g|)$
   \STATE $\delta \leftarrow \delta+\alpha^{\prime} g$
   %\STATE $\tilde{x} = \tilde{x} + \alpha^{'}gradients$
   \ENDFOR
   \STATE
   \STATE {\bfseries Return:} $\delta$
\end{algorithmic}
\end{algorithm}
\begin{figure}[t]
\vskip 0.05in
\begin{center}
\centerline{\includegraphics[width=\columnwidth]{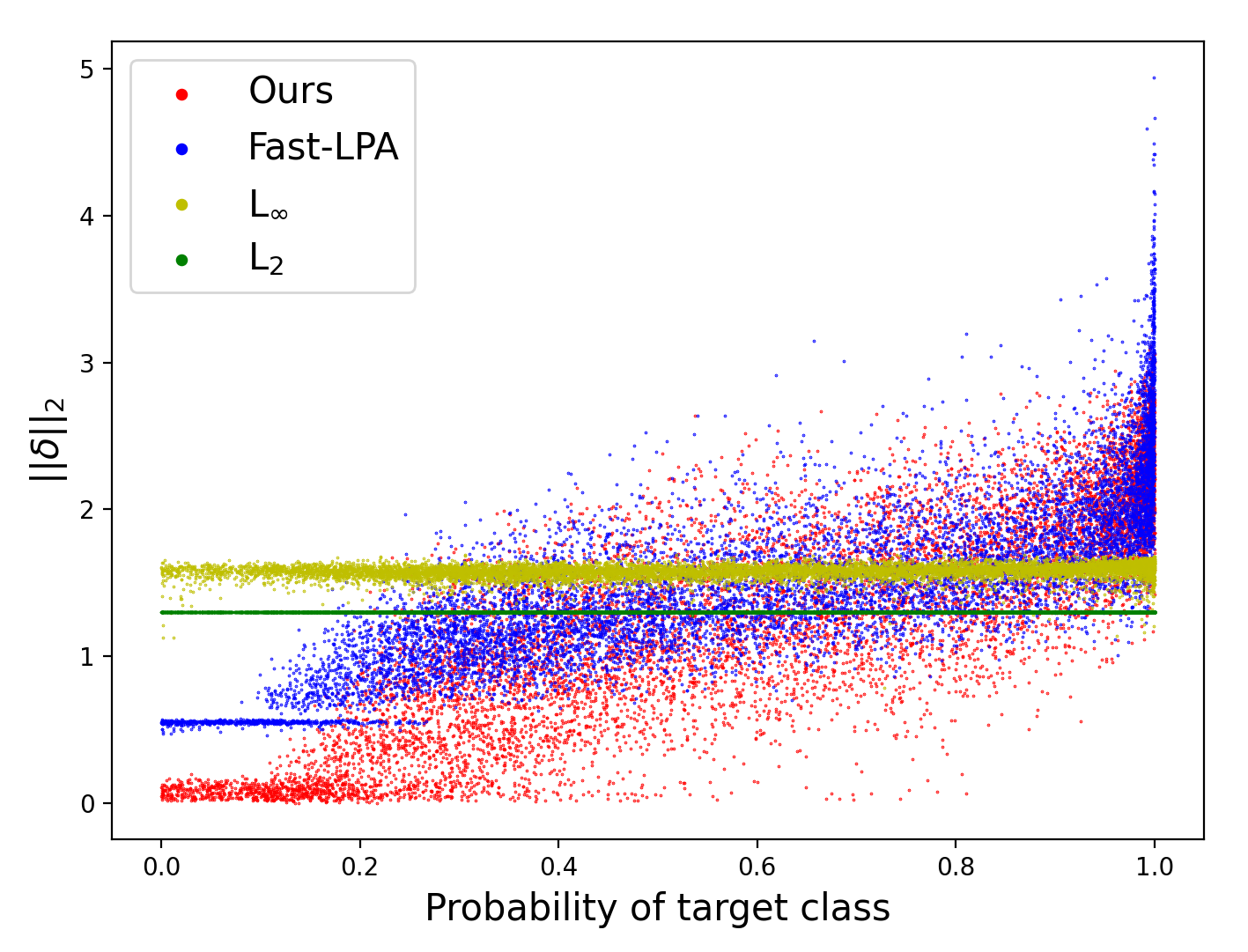}}
\vskip 0.05in
\caption{$\ell_2$ norm of perturbations generated using our attack, PGD ($\ell_2$, $\ell_\infty$) and Fast-LPA vs. probability of the correct class that is estimated by the network for clean examples in the CIFAR-10 test dataset.}
\label{scatter}
\end{center}
\vskip -0.3in
\end{figure}
%\newline Now we have a general adversarial attack that avoidable constraints have been detached from it as much as possible so it can model features of other attacks and possible attacks that may be proposed in the future.

Next, using our proposed attack $\delta_{\text{Lag.}, i}$, we perform adversarial training in the form of:
\[\min_\theta \sum_i \ell(f_\theta(x_i + \delta_{\text{Lag.}, i}), y_i)\]
%\[\min_\theta \sum_i \max_{\delta \in \Delta} \ell(f_\theta(x_i + \delta), y_i),\]
%where we used cross-entropy error function as \emph{loss}
and we use SGD to minimize the total loss. %Since we are using an attack that can model disparate features, we expect our model to learn better features and reach better robustness against unseen attacks.
Note that there could be a trade-off between standard accuracy and robust accuracy \cite{tsipras2019}. Therefore, by decreasing $\lambda$ and allowing larger perturbation sizes, one would expect a lower clean accuracy. For this reason, the training attack parameters 
%our standard accuracy will decrease so parameters
should be adjusted in a way that model reaches a satisfactory standard accuracy.
%\newline In addition to eliminating avoidable constraints, 
\subsection{Relation to Earlier Work}

Here, we discuss the connection between our method and some relevant earlier work such as MART \cite{Wang2020Improving} and instance adaptive method \cite{Balaji2019}. MART investigated the effect of misclassified and correctly classified examples on the robustness of adversarially trained models. They suggested that misclassified examples and correctly classified examples should be treated differently during training. Instance adaptive method increased the perturbation margin for each sample specifically until the sample is correctly classified. They trained the network using the clean loss for misclassified examples, and adversarial loss for correctly classified examples. 

Our work is a generalization of these methods. Our proposed attack uses a constant Lagrange multiplier that causes larger perturbations for confidently classified examples and approximately no perturbation for examples that are confidently misclassified. To empirically demonstrate this point, we plotted $\| \delta \|_2$ against the probability of belonging to the correct class using the classifier output, for each sample in the CIFAR-10 test dataset. Here, we considered $\delta$ be obtained using PGD ($\ell_2$, $\ell_\infty$) \cite{Madry2018}, $\delta_{\text{Lag.}}$ (ours), and Fast-LPA used in PAT \cite{Laidlaw2020}. Fig. \ref{scatter} shows the plot. Based on this figure, $\ell_2$ norm of perturbations for PGD attacks are approximately constant, but in our attack and Fast-LPA that are using the Lagrangian formulation, it has direct relationship with classifier confidence of correctly classifying the example. The advantage of this adaptive perturbation magnitude is discussed in \cite{Balaji2019}.

%However, it shows that our attack almost does not change the examples that are most likely classified incorrectly.  %However, this is not exactly the case in Fast-LPA,
%as $\text{LPIPS}(x, x+\delta) \approx 0$ does not necessarily imply $\| \delta \|_2 \approx 0$.

\subsection{Theoretical Insights}\label{theoretical}
Here, we provide some theoretical insights that support the Lagrangian formulation. Let
\[ U_\Delta := \max_{\delta \in \Delta} \ell(f_\theta(x + \delta), y), \]
which is the maximum loss at the input $x$ for a custom threat model $\Delta$. We let  $\delta_\Delta$ denote the perturbation that results in $U_\Delta$.
Also, let
\[L_2^\epsilon := \max_\delta \ \ell(f_\theta (x+\delta), y) \ \ \ s.t. \ \ \ {\| \delta \|}_2 \leq \epsilon.\]
%and $\delta^\star$ be the optimal solution for it. 
As mentioned above, we use the Lagrangian to solve this problem that gives us
\[\delta_{\text{Lag.}}^\star := \argmax_\delta \ \ell(f_\theta (x+\delta), y) -\lambda {\| \delta \|}_2.\]
We next define $\Delta \epsilon$ as:
\[\Delta \epsilon := |{\| \delta_\Delta \|}_2 -{\| \delta_{\text{Lag.}}^\star \|}_2| \ \]
On the other hand, using the envelope theorem \cite{Milgrom2002}, we have:
\[\dfrac{\partial L^\epsilon_2}{\partial \epsilon} = \lambda,\]
for the $\epsilon$ that corresponds to $\lambda$ that is the constant Lagrange multiplier in our method. Now, using the Taylor series, we  have:
\[ U_\Delta \leq L_2^{{\| \delta_{\Delta} \|}_2 } \leq L_2^{{\| \delta^\star_{\text{Lag.}} \|}_2 } + \lambda .(\Delta \epsilon) + o(\Delta \epsilon)\]
So if $\Delta \epsilon$ is small, knowing that $\lambda$ is fixed, bounding $L_2^{{\| \delta_{\text{Lag.}}^\star \|}_2 }$ using adversarial training with our proposed attack would be equivalent to bounding $U_\Delta$ that is related to the attack model $\Delta$ that has slightly different $\ell_2$ norm than what is considered in the training. In section \ref{Effective_lam}, this proof is discussed more practically.

\begin{table*}[t]
  \begin{small}
  \begin{center}
  \caption{Test robust accuracy of our model along with other models against training attacks and unseen attacks in the CIFAR-10 dataset. Clean accuracy and robust accuracy of our model and other state-of-the-art models against LPA (bound=0.5, steps=200), $\ell_\infty$(bound=8/255, steps=40),  $\ell_2$(bound=1.0, steps=40), RecolorAdv(bound=0.06, steps=100), StAdv(bound=0.05, steps=100), Elastic(bound=0.5, steps=50), JPEG(bound=0.125, steps=50), Guassian noise (mean=0.0, variance=0.05), Gaussian Blur (kernel size=5, sigma=1.5), and PGD$_0$(bound=10, steps=20) are reported in this table. Unseen mean shows average accuracy against unseen attacks and union mean shows average accuracy against all the attacks.}
  \vskip 0.15in
  \setlength{\tabcolsep}{3.5pt} % Default value: 6pt
  \renewcommand{\arraystretch}{1.3} % Default value: 1
    \begin{tabular}{c|c|ccc|ccccccc|cc}
    \multicolumn{1}{c|}{\multirow{1}[3]{*}{Training}} & \multirow{2}[3]{*}{Clean} & \multicolumn{3}{c|}{Training attacks} & \multicolumn{7}{c|}{Unseen attacks}                                          & \multirow{1}[3]{*}{Unseen} & \multirow{1}[3]{*}{Union} \\
\cmidrule{3-12}       method   &       & LPA   & $\ell_\infty$  & $\ell_2$    & Recolor & StAdv & Elastic & Jpeg  & Noise & Blur & PGD$_0$  &    mean   & mean \\
    \midrule
    Trades $\ell_\infty$ & 79.54 & 0.18  & 51.91 & 38.66 & 60.57 & 24.22 & 37.56 & 36.12 & 47.70 & 56.45 & 50.91 & 44.79 & 40.43 \\
    MART $\ell_\infty$ & 80.01 & 0.29  & \textbf{54.03} & 38.32 & 59.61 & 18.45 & 38.63 & 36.87 & 42.23 & 54.85 & 50.33 & 43.00 & 39.36 \\
    MSD   & 80.62 & 1.69  & 48.69 & 49.11 & 55.34 & 14.86 & 33.88 & 53.86 & 42.56 & 57.71 & \textbf{70.19} & 46.91 & 42.79 \\
    DDN $\ell_2$ &  72.18 & \textbf{17.92} & 42.44 & 46.34 & 29.06 & 3.65  & 35.36 & 34.44 & 51.32 & 46.23 & 52.67 & 36.10 & 35.94\\
    MMA $\ell_2$ & 82.07 & 2.67  & 47.74 & 50.79 & 49.97 & 27.74 & 34.57 & 55.20  & 39.67 & \textbf{63.66} & 45.32 & 45.16 & 41.73\\
    IAAT $\ell_\infty$ & \textbf{82.53} & 14.97 & 47.89 & 38.34 & 39.79 & 2.74  & \textbf{42.98} & 35.84 & \textbf{69.29} & 60.55 & 45.83 & 42.43 & 39.82 \\
    AT $\ell_2$ & 78.59 & 1.20   & 45.60  & 51.48 & 53.57 & 19.73 & 26.67 & 60.14 & 52.30 & 58.35 & 63.30  & 47.72 & 43.23 \\
    AT $\ell_2$ ES & 74.90  & 1.97  & 47.88 & 53.38 & 57.98 & 24.87 & 33.66 & 58.48 & 49.74 & 56.62 & 60.06 & 48.77 & 44.46\\
    Threshold $\ell_2$ & 78.08 & 1.17  & 43.64 & 50.38 & 56.09 & 25.93 & 26.18 & 57.33 & 52.21 & 59.11 & 63.10  & 48.56 & 43.51 \\
    AT $\ell_\infty$ & 79.98 & 0.30   & 53.41 & 38.45 & 58.94 & 16.17 & 34.28 & 38.57 & 36.61 & 54.08 & 47.25 & 40.84 & 37.81 \\
    AT $\ell_\infty$ ES & 79.36 & 0.42  & 53.31 & 38.77 & 58.98 & 17.62 & 33.76 & 39.21 & 40.72 & 54.13 & 49.25 & 41.95 & 38.62 \\
    Threshold $\ell_\infty$ & 80.90 & 0.29 & 51.45 & 37.69 & 62.59 & 22.31 & 38.26 & 33.8 & 50.08 & 56.62 & 47.98 & 44.52 & 40.11\\
    PAT   & 80.31 & 5.14  & 43.19 & 46.19 & 66.45 & \textbf{45.31} & 34.02 & 48.16 & 39.06 & 55.89 & 57.22 & 49.44 & 44.06 \\
    \textcolor[rgb]{ .122,  .306,  .471}{\textit{\textbf{Ours}}}  & 80.43 & 1.50  & 48.12 & \textbf{54.17} & \textbf{69.14} & 41.57 & 36.83 & \textbf{61.37} & 52.93 & 62.36 & 63.37 & \textcolor[rgb]{ .122,  .306,  .471}{\textit{\textbf{55.37}}} & \textcolor[rgb]{ .122,  .306,  .471}{\textit{\textbf{49.14}}} \\
    \end{tabular}%
  \label{CIFAR10 results}
  \end{center}
  \end{small}
  \vskip -0.1in
\end{table*}%
\begin{table*}[t]
  \begin{small}
  \begin{center}
  \caption{Test robust accuracy of our model along with other models against training attacks and unseen attacks in the ImageNet-100 dataset. Clean accuracy and robust accuracy of our model and other state-of-the-art models against LPA (bound=0.25, steps=200), $\ell_\infty$ (bound=4/255, steps=40),  $\ell_2$(bound=4.0, steps=40), RecolorAdv(bound=0.06, steps=100), StAdv(bound=0.05, steps=100), Elastic(bound=0.5, steps=50), JPEG(bound=0.125, steps=50), Fog(bound=256, steps=50), Gabor(bound=0.125, steps=50), Snow(bound=0.125, steps=50), Guassian noise (mean=0.0, variance=0.05), and Gaussian Blur (kernel size=5, sigma=1.5) are reported in this table. Unseen mean shows average accuracy against unseen attacks and union mean shows average accuracy against all the attacks.}
  \vskip 0.15in
  \setlength{\tabcolsep}{3.5pt} % Default value: 6pt
  \renewcommand{\arraystretch}{1.3} % Default value: 1
    \begin{tabular}{c|c|cccc|cccccccc|rr}
    \multicolumn{1}{c|}{\multirow{1}[3]{*}{Training}} & \multirow{2}[3]{*}{Clean} & \multicolumn{4}{c|}{Training attacks} & \multicolumn{8}{c|}{Unseen attacks}                   & \multicolumn{1}{c}{\multirow{1}[3]{*}{Unseen}} & \multicolumn{1}{c}{\multirow{1}[3]{*}{Union}} \\
\cmidrule{3-14}       method   &       & LPA   & $\ell_\infty$  & $\ell_2$    & JPEG  & Recolor & StAdv & Elastic & Fog   & Gabor & Snow  & Noise & Blur &   mean    & mean \\
    \midrule
    AT $\ell_2$ & 70.14 & 2.30 & 40.54 & 40.07 & 46.72 & 21.67 & 16.58 & 40.70  & \textbf{17.54} & 50.62 & 34.7  & 25.06 & 60.67 & 33.44 & 33.10 \\
    AT $\ell_\infty$ & 70.28 & 0.02  & \textbf{50.35} & 13.24 & 7.74  & 32.51 & 14.41 & 46.38 & 12.96 & \textbf{55.20}  & \textbf{45.36} & 8.35  & 58.60  & 34.22 & 28.76 \\
    JPEG  & \textbf{72.43} & 0.53  & 46.21 & 35.88 & \textbf{65.12} & \textbf{37.67} & 8.61  & 44.44 & 14.46 & 54.60  & 36.44 & \textbf{31.93} & \textbf{64.72} & 36.61 & 36.72 \\
    PAT   & 70.11 & 6.42  & 43.14 & 43.29 & 49.28 & 32.10 & 34.54 & 46.92 & 16.17 & 52.92 & 39.36 & 18.66 & 58.09 & 37.35 & 36.74 \\
     \textcolor[rgb]{ .122,  .306,  .471}{\textit{\textbf{Ours}}}  & 69.95 & \textbf{8.93}  & 46.47 & \textbf{46.59} & 52.16 & 34.32 & \textbf{35.26} & \textbf{47.86} & 17.16 & 54.24 & 42.84 & 29.43 & 62.95 &  \textcolor[rgb]{ .122,  .306,  .471}{\textit{\textbf{40.51}}} &  \textcolor[rgb]{ .122,  .306,  .471}{\textit{\textbf{39.85}}}\\
    \end{tabular}%

  \label{ImageNet results}%
  \end{center}
  \end{small}
  \vskip -0.1in
\end{table*}%
\section{Experiment}\label{Experiment}
\subsection{Setup}
To this end, we proposed a new adversarial attack for training time which aims to be robust against unseen threat models. To evaluate our method, we adversarially trained the ResNet-18 model using our attack on CIFAR-10 \cite{Krizhevsky2009}, and the ResNet-34 on the ImageNet-100 dataset (100-class subset of ImageNet \cite{ILSVRC152015} containing every 10th class by the WordNet ID). We compare the robust accuracy of our method with other state-of-the-art methods against attacks that are used during training and a number of unseen attacks for each of these datasets. For training, we used standard techniques such as learning rate decay and warm-up to make the convergence faster. More specifically, following \cite{Laidlaw2020}, the learning rate for weight updates in SGD is set to $0.1$ initially, and reduced by a factor of $0.1$ at epochs 70 and 90 in CIFAR-10. For the ImageNet-100, it is initially set to 0.1 and reduced by a factor of 0.1 at epochs 30, 60, and 80. The models are trained for 100 and 90 epochs in CIFAR-10 and ImageNet-100, respectively. The warm-up step is pre-training the networks with clean data samples for three epochs. We applied the same techniques in training  rest of the models to have a fair comparison. In order to make the robust accuracy of different models comparable, we tuned the attack hyperparameters in each method to get similar/same clean accuracy. This has been achieved by changing the  bounds on the perturbation size of each method such that we reach a similar standard accuracy for all the models.

\subsection{Baseline Methods} 
We compare our method with the best methods that have reached acceptable and competitive robust accuracies. For CIFAR-10 dataset, we use TRADES \emph{\((\ell_\infty)\)} \cite{Zhang2019} as a method that considered the trade-off between standard and robust accuracy, and MART \cite{Wang2020Improving} since it treats misclassified examples differently and adversarial training (AT) with PGD \emph{\((\ell_\infty, \ell_2)\)} \cite{Madry2018} with or without early stopping (ES) . We also compare with adaptive methods such as IAAT \cite{Balaji2019}, MMA \cite{Ding2020}, and DDN \cite{Rony2019}. In addition, MSD \cite{Miani2020}, and PAT \cite{Laidlaw2020} tried to reach robustness against multiple perturbations and perceptual attacks that we  also compare against. All methods have been used with similar settings introduced for that method. Only in some cases for a fairer comparison, the perturbation budget size of training has been increased to achieve a better robust accuracy at the cost of reducing clean accuracy. For example, we trained the AT-$\ell_2$ model with $\epsilon=1.3$ and AT-$\ell_\infty$ with $\epsilon=10/255$.

Among the mentioned cases, only adversarial training with PGD and Fast-LPA (PAT method) have been examined with ImageNet dataset, due to shortage of computational resources. For a better comparison, we also consider adversarial training with JPEG attack \cite{Kang2020}. Prior work \cite{Kang2020, Laidlaw2020} have demonstrated that the models that are trained using the JPEG attack, as opposed to other attacks such as Elastic, Fog, Gabor, Snow, RecolorAdv, and StAdv, have better generalization against unseen attacks. We also do not use training with the union of attacks for two reasons. First, \cite{Laidlaw2020} demonstrated that the PAT method is more robust than optimizing the model with a random attack from the union of attacks or the average or worst case loss. Second, it is inconsistent with our goal of examining the final model with unseen attacks since this method assumes that all types of possible attacks are known.

However, to examine whether the instance adaptive perturbation bound has played a role in improving the results, as demonstrated in Fig. \ref{scatter} and discussed in section \ref{proposed}, we train a model using PGD-$\ell_\infty$ and another using PGD-$\ell_2$ with different perturbation bounds based on the probability of correct class given the clean examples. For this goal, we set the thresholds \{0.1, 0.25, 0.5\} to divide [0, 1] interval for probability of the correct class into 4 sub-intervals, and set the maximum allowable perturbation in these sub-intervals to 0.03$\epsilon$, 0.3$\epsilon$, 0.55$\epsilon$, and $\epsilon$, respectively. We name these methods Threshold-$\ell_\infty$/$\ell_2$ and we test them only on CIFAR-10 dataset, for the sake of performing an ablation study. 

\subsection{Unforeseen Attacks} 
To evaluate models that are trained on CIFAR-10, we use the attacks that are employed in the training time (PGD\emph{\((\ell_\infty, \ell_2)\)}, LPA), and unseen attacks that are widely used in earlier work. \cite{Kang2020} suggested 5 different types of attacks, including Elastic, JPEG, FOG, Gabor, and Snow, as unseen attacks and calibrated maximum distortion size for some of them in CIFAR-10, and all of them in the ImageNet-100 dataset. At the test time, we  use the calibrated ones with maximum distortion size of \emph{\(\epsilon_3\)} for CIFAR-10, and \emph{\(\epsilon_2\)} for ImageNet-100 based on \cite{Kang2020} calibration. We also use StAdv \cite{xiao2018}, RecolorAdv \cite{Laidlaw2019}, PGD$_0$ \cite{Croce2019}, Gaussian noise, and Gaussian blurring to make unseen attacks more comprehensive. These attacks are selected to cover a diversity of corruptions. For the ImageNet-100 dataset, we use the same attacks that are mentioned above except PGD$_0$ since it has not been examined on the ImageNet. Note that attack parameters are selected following parameters that are widely used in earlier work or instructions of the paper that proposed the attack. Detailed information about attacks are mentioned in the caption of Tables \ref{CIFAR10 results} and \ref{ImageNet results}.

\subsection{Adversarial Accuracy} \label{adv_acc}

Results of evaluating our method and other mentioned methods against training and unseen attacks are listed in Table \ref{CIFAR10 results} for CIFAR-10, and Table \ref{ImageNet results} for the ImageNet-100 dataset. These results demonstrate that our method is  more robust than others against unseen attacks and union of unseen and training attacks on average. In CIFAR-10, our model has 5.93\% and 4.68\% higher average robust accuracy than others against unseen attacks and union of attacks, respectively. In ImageNet-100, our model also reaches 3.16\% and 3.11\% higher robustness accuracy against unseen attacks, and union of attacks, respectively.

We note that (Threshold-$\ell_\infty$) and (Threshold-$\ell_2$)  have higher robustness than PGD-$\ell_\infty$ and PGD-$\ell_2$, respectively, that demonstrates the influence of lower perturbation margin for misclassified examples on unseen attack generalization. As explained in section \ref{proposed}, our Lagrangian method with a fixed multiplier generally does the same thing.

\subsection{Computational Efficiency}

Our proposed attack is also computationally efficient. Since the comparison of execution time depends on the exact GPU/CPU that is used for training, mentioning numbers for the training time does not help much in practice. Therefore, we only compare the execution time order. Each iteration of our attack is approximately equal to one iteration in a PGD attack in terms of the running time. Considering that we use only 5 iterations, the execution time of our attack is almost equivalent to the PGD-5 attack.

We note that PGD is usually used with 10 or more iterations at the training time to achieve robustness against perturbations with larger $\ell_p$-radii \cite{andriushchenko2020}. So our attack takes about half the time required for the conventional PGDs. Fast-LPA attack that is used in PAT, generates perturbation in 10 iterations, while we use only 5. However, it also computes the LPIPS distance in each iteration that makes each iteration more time-consuming, about 1.2 times for Fast-LPA with AlexNet. So our attack is also much faster than the Fast-LPA. Other attacks such as JPEG, are also computationally worse than ours, since they use more iterations or each iteration of their method consumes more time than ours. This makes our method scalable for training large models on large datasets.

\subsection{Remarks}

{\bf Effective $\lambda$}: \label{Effective_lam} \newline
To validate our theoretical insights furthermore, we estimated the effective average $\lambda$, which is defined as the derivative of  adversarial loss with respect to $\epsilon$, for some of the CIFAR-10 models that are investigated in Table \ref{CIFAR10 results}. Based on the results in Table \ref{tab:effective_lam}, our method has the lowest such $\lambda$ and based on our insights in section \ref{theoretical}, lower $\lambda$ causes a tighter upper bound on the loss from unforeseen attacks and improves the accuracy against them, which is consistent with our experiments in section \ref{adv_acc}.

\begin{table}[t]
  \centering
  \begin{small}
  \caption{Average and standard deviation of effective $\lambda$ in models trained with CIFAR-10 for samples in this dataset.}
  \vskip 0.1in
  \setlength{\tabcolsep}{6.0pt} % Default value: 6pt
  \renewcommand{\arraystretch}{1.2} % Default value: 1
    \begin{tabular}{c|cc}
    model & Avg. $\lambda$ & Std. $\lambda$ \\
    \midrule
    DDN $\ell_2$   & 1.14  & 0.84 \\
    MMA $\ell_2$  & 1.65  & 1.72 \\
    IAAT $\ell_\infty$ & 0.96  & 1.99 \\
    AT $\ell_2$ & 0.86  & 0.87 \\
    AT $\ell_\infty$ & 0.63  & 0.54 \\
    PAT   & 0.76  & 0.9 \\
    \textbf{Ours} & \textbf{0.38} & \textbf{0.43} \\
    \end{tabular}%
  \label{tab:effective_lam}%
  \end{small}
\end{table}%

{\bf Signed vs. pure input gradients}: \newline
We note that using the pure input gradients, as opposed to the signed gradient, in the iterative updates, would help to craft stronger perturbations even in the $\ell_\infty$ threat model.
To demonstrate this, we test $\ell_2$ and $\ell_\infty$ adversarially trained models against PGD-40 ($\ell_\infty$) and a modified version of PGD-40 (MPGD-40) that does not take the sign from gradients but normalizes them between -1 and 1 to bring gradients from different samples in the same scale. Results listed in Table \ref{gradient sign} demonstrate that this small change has made the attack stronger and confirms our claim.
\begin{table}[t]
  \begin{center}
  \begin{small}
  \caption{Testing $\ell_2$ and $\ell_\infty$ adversarially trained models against $\ell_\infty$ attacks with small and large bounds, using 40 iteration for optimizing the perturbations. In CIFAR-10, we set small=8/255 and large=16/255 and in ImageNet-100, we set small=4/255 and large=8/255. MPGD is our modified version of PGD without using the gradient sign. Step size for each method is selected with a grid search.}
  \vskip 0.15in
  \setlength{\tabcolsep}{3.5pt} % Default value: 6pt
  \renewcommand{\arraystretch}{1.3} % Default value: 1
  \begin{tabular}{c|l|cc|cc}
    \multirow{2}[1]{*}{Dataset} & \multicolumn{1}{c|}{\multirow{2}[1]{*}{Model}} & PGD   & MPGD  & PGD   & MPGD \\
          &       & (small) & (small) & (large) & (large) \\
    \midrule
    \multirow{2}[2]{*}{CIFAR-10} & AT $\ell_2$ & 45.6  & 43.05 & 17.95 & 10.62 \\
          & AT $\ell_\infty$ & 53.41 & 52.06 & 27.19 & 24.01 \\
    \midrule
    \multirow{2}[1]{*}{ImageNet-100} & AT $\ell_2$ & 40.54 & 38.91 & 17.56 & 15.13 \\
          & AT $\ell_\infty$ & 50.5  & 49.84 & 31.15 & 27.76 \\
    \end{tabular}%
  \label{gradient sign}%
  \end{small}
  \end{center}
  \vskip -0.15in
\end{table}
\begin{figure*}[t]
\begin{center}
\includegraphics[width=\textwidth]{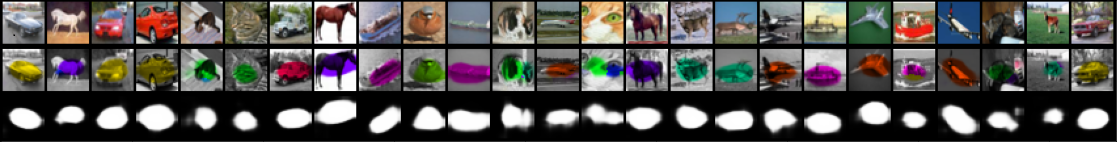}
\vskip 0.05in
\caption[width=\textwidth]{Segmentation results for CIFAR-10. First row shows some images from the CIFAR-10 dataset and the next rows is those images after separating background from the foreground. Last row also shows the segmentation mask for the images.}
\label{Cifar segmentation}
\end{center}
\vskip -0.15in
\end{figure*}
\begin{figure*}[t]
\begin{center}
\includegraphics[width=\textwidth]{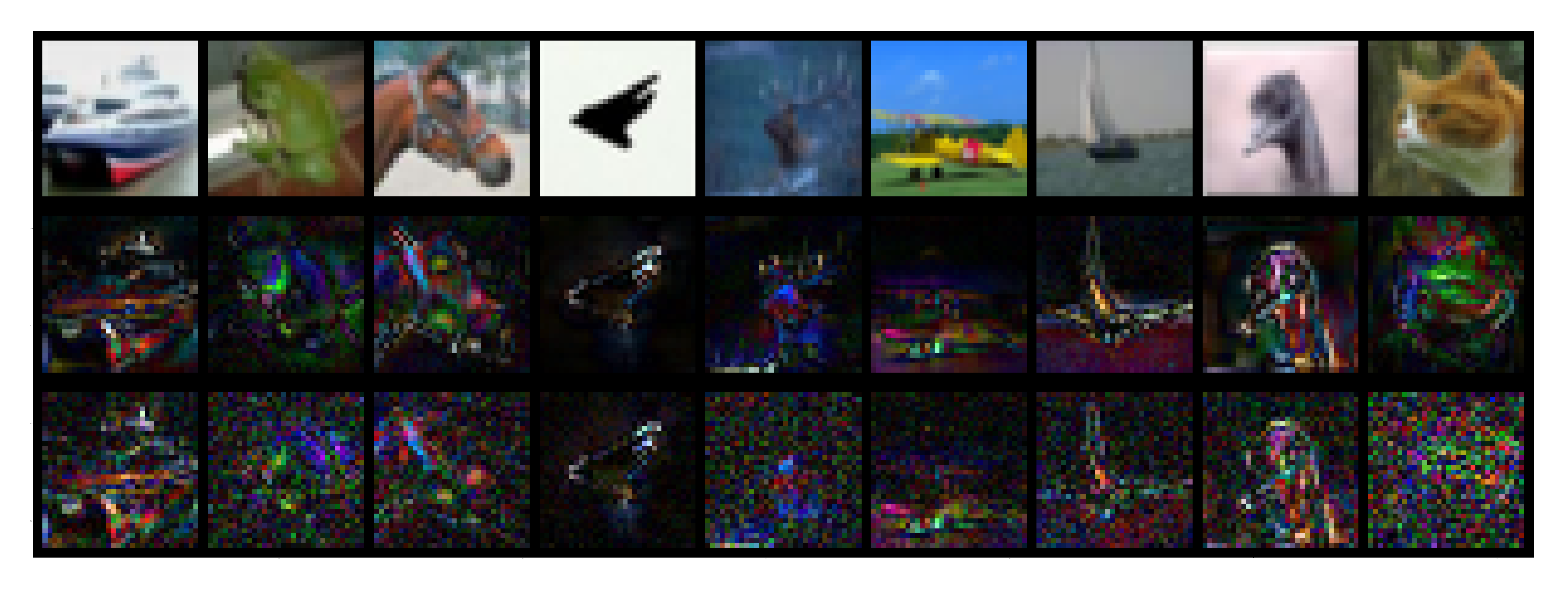}
\vskip -0.05in
\caption[width=\textwidth]{Comparison of the generated perturbations using LPIPS distance, and \emph{\({\| \delta \|}_2\)} as the penalty term. First row shows some images from CIFAR-10 dataset,  second row shows generated perturbation for images in first row using \emph{\({\| \delta \|}_2\)} as the distance metric, and the third row is the generated perturbation using the LPIPS distance.}
\label{LPIPS VS L2}
\end{center}
\vskip -0.10in
\end{figure*}

{\bf LPA bias towards the background}: \newline
Next, we notice that the LPA attack is more biased to perturb pixels in the background of the input image.
Here, we practically demonstrate that using LPIPS instead of \emph{\({\| \delta \|}_2\)} makes perturbations to be focused more on the background of the image. To validate this claim, we use segmentation as a tool to separate the background and foreground of the input image. We then calculate the ratio of total perturbations in the foreground to total perturbations in the background. For this goal, we used the deep networks with the same architecture to \cite{SegmentCifar} for CIFAR-10, and \cite{Zagoruyko2017AT} for ImageNet-100 to separate background and foreground in clean images. This gives us an estimation of the probability of a pixel being part of the main object.
Segmentation results for CIFAR-10 are shown in Fig. \ref{Cifar segmentation}, which shows a decent performance in spite of low resolution images.

Using the segmentation results, we define F2B as a measure to determine the ratio of total perturbations in the foreground to total perturbations in the background as:
\[F2B(\delta) := \dfrac{\sum_i (|\delta_i| \times p_i)/\sum_i p_i}{\sum_i(|\delta_i| \times (1-p_i))/\sum_i (1-p_i)},\]
where $p_i$ is the calculated probability of pixel $i$ belonging to the foreground, and $\delta_i$ is the perturbation value at the same pixel. For CIFAR-10, average F2B is 1.42 for LPIPS, while it is 1.69 for \emph{\({\| \delta \|}_2\)}. For ImageNet-100, it is 1.06 and 1.16, respectively. This confirms higher focus of the LPA perturbation in the background than the $\ell_2$ norm. To make this more clear, output perturbation of these attacks are shown in Fig. \ref{LPIPS VS L2} for some sample images from the CIFAR-10 dataset. It is obvious that generated perturbations with LPIPS as the distance metric has made more changes in the background than generated perturbations with the $\ell_2$ norm. 

\begin{figure}[t!]
\begin{center}
\centerline{\includegraphics[width=\columnwidth]{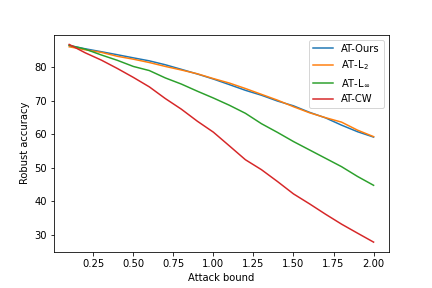}}
%\vskip 0.1in
\caption{Comparison of robust accuracy for C\&W-trained model with three other methods against PGD-$\ell_2$ on three first classes of CIFAR-10 dataset. PGD-$\ell_2$ bound is gradually increased from 0.1 to 2.}
\label{CW}
\end{center}
\vskip -0.2in
\end{figure}

{\bf C\&W attack for adversarial training}: \newline
As our proposed attack is similar to the C\&W attack, we also investigate adversarially trained model with the C\&W attack. Since C\&W is extremely time-consuming, we use only the first three classes of the CIFAR-10 dataset. We evaluate the resulting models against PGD-$\ell_2$ with various bounds. Based on the Fig. \ref{CW}, robust accuracy of the C\&W-trained model decreases more than other models as the allowable range of perturbation increases, due to training of the model with minimally perturbed examples.

{\bf Comparison against other perturbation budgets}: \newline
Here, we examine effect of the test time perturbation budget on results that are presented in Table \ref{CIFAR10 results} and \ref{ImageNet results}. Although we selected the perturbation bounds according to the prior work in these experiments, we evaluated against other perturbation sizes to give a more comprehensive evaluation of the method. For this purpose, we evaluated a number of methods from Table \ref{CIFAR10 results} against PGD attacks with different bounds. Results are shown in Table \ref{attack bound}. It shows that if one method is more resistant to an attack with a specific bound than the others, it would also be more resistant against that attack with other bounds. Therefore, this parameter does not have much effect on the final results.
\begin{table}[t]
  \centering
  \begin{small}
  %\vskip 0.05in
  \caption{Evaluating five different methods trained on CIFAR-10 against PGD attacks with three different bounds.}
  \vskip 0.15in
   \setlength{\tabcolsep}{4.5pt} % Default value: 6pt
  \renewcommand{\arraystretch}{1.3} % Default value: 1
    \begin{tabular}{c|ccc|ccc}
    Training & \multicolumn{3}{c|}{$\ell_2$} & \multicolumn{3}{c}{$\ell_\infty$} \\
\cmidrule{2-7}    method & 0.75  & 1.00     & 1.25  & 6/255 & 8/255 & 10/255 \\
    \midrule
    MSD   & 58.43 & 49.11 & 41.48 & 57.13 & 48.69 & 39.97 \\
    AT $\ell_2$ & 59.75 & 51.48 & 45.05 & 54.37 & 45.60  & 37.48 \\
    AT $\ell_\infty$ & 51.49 & 38.45 & 31.47 & 60.84 & 53.41 & 45.83 \\
    PAT   & 55.55 & 46.19 & 40.67 & 51.86 & 43.19 & 34.62 \\
    Ours  & 62.39 & 54.17 & 46.52 & 57.81 & 48.12 & 38.84 \\
    \end{tabular}%
  \label{attack bound}%
  \end{small}
  %\vskip -0.1in
\end{table}%

\section{Conclusion}

Previous works generally examine the robustness only against existing attacks and do not pay attention to unforeseen threat models. This creates a rivalry between defenses and attacks, and for every new defense, a new successful attack emerges, and vice versa. In doing so, we tried to solve this problem and came up with a method that would make the classifier more resistant to unforeseen attacks. To this end, we introduced a new attack that does not have the limitations and biases of previous attacks at the training time and determines the perturbation margin associated with each sample according to the sample itself, using the Lagrangian penalty. Moreover, our introduced method is much faster than others and allows its use for large datasets. Finally, we demonstrated the effectiveness of our method in improving generalization in the CIFAR-10 and ImageNet-100 datasets against many unseen attacks. Our work has shown that using attacks that perform well during test time is not sufficient for training network against unforeseen attacks and it is necessary to design a special teacher for the training time that, in addition to teaching a general case of possible perturbations to the classifier, allows the teacher to determine the allowable amount of perturbation budget for each sample according to the sample itself.

\bibliography{example_paper}

\begin{thebibliography}{42}
\providecommand{\natexlab}[1]{#1}
\providecommand{\url}[1]{\texttt{#1}}
\expandafter\ifx\csname urlstyle\endcsname\relax
  \providecommand{\doi}[1]{doi: #1}\else
  \providecommand{\doi}{doi: \begingroup \urlstyle{rm}\Url}\fi

\bibitem[Akhtar \& Mian(2018)Akhtar and Mian]{Akhtar2018}
Akhtar, N. and Mian, A.
\newblock Threat of adversarial attacks on deep learning in computer vision: A
  survey.
\newblock \emph{arXiv preprint arXiv:1801.00553}, 2018.

\bibitem[Andriushchenko \& Flammarion(2020)Andriushchenko and
  Flammarion]{andriushchenko2020}
Andriushchenko, M. and Flammarion, N.
\newblock Understanding and improving fast adversarial training.
\newblock In \emph{Advances in Neural Information Processing Systems
  (NeurIPS)}, 2020.

\bibitem[Athalye et~al.(2018{\natexlab{a}})Athalye, Carlini, and
  Wagner]{Athalye2018}
Athalye, A., Carlini, N., and Wagner, D.~A.
\newblock Obfuscated gradients give a false sense of security: Circumventing
  defenses to adversarial examples.
\newblock In \emph{International Conference on Machine Learning (ICML)},
  2018{\natexlab{a}}.

\bibitem[Athalye et~al.(2018{\natexlab{b}})Athalye, Engstrom, Ilyas, and
  Kwok]{Athalye2018b}
Athalye, A., Engstrom, L., Ilyas, A., and Kwok, K.
\newblock Synthesizing robust adversarial examples.
\newblock In \emph{International Conference on Machine Learning (ICML)},
  2018{\natexlab{b}}.

\bibitem[Balaji et~al.(2019)Balaji, Goldstein, and Hoffman]{Balaji2019}
Balaji, Y., Goldstein, T., and Hoffman, J.
\newblock Instance adaptive adversarial training: Improved accuracy tradeoffs
  in neural nets.
\newblock \emph{arXiv preprint arXiv:1910.08051}, 2019.

\bibitem[Carlini \& Wagner(2017)Carlini and Wagner]{Carlini2017}
Carlini, N. and Wagner, D.
\newblock Towards evaluating the robustness of neural networks.
\newblock In \emph{IEEE Symposium on Security and Privacy (SP)}, 2017.

\bibitem[Chen et~al.(2020)Chen, Liu, Chang, Cheng, Amini, and Wang]{Chen2020}
Chen, T., Liu, S., Chang, S., Cheng, Y., Amini, L., and Wang, Z.
\newblock Adversarial robustness: From self-supervised pre-training to
  fine-tuning.
\newblock In \emph{IEEE Conference on Computer Vision and Pattern Recognition
  (CVPR)}, 2020.

\bibitem[Croce \& Hein(2019)Croce and Hein]{Croce2019}
Croce, F. and Hein, M.
\newblock Sparse and imperceivable adversarial attacks.
\newblock In \emph{International Conference on Computer Vision (ICCV)}, 2019.

\bibitem[Croce \& Hein(2020)Croce and Hein]{croce2020}
Croce, F. and Hein, M.
\newblock Reliable evaluation of adversarial robustness with an ensemble of
  diverse parameter-free attacks.
\newblock In \emph{International Conference on Machine Learning (ICML)}, 2020.

\bibitem[Ding et~al.(2020)Ding, Sharma, Lui, and Huang]{Ding2020}
Ding, G.~W., Sharma, Y., Lui, K. Y.~C., and Huang, R.
\newblock {MMA} training: Direct input space margin maximization through
  adversarial training.
\newblock In \emph{International Conference on Learning Representations
  (ICLR)}, 2020.

\bibitem[Dong et~al.(2018)Dong, Liao, Pang, Su, Zhu, Hu, and Li]{Dong2018}
Dong, Y., Liao, F., Pang, T., Su, H., Zhu, J., Hu, X., and Li, J.
\newblock Boosting adversarial attackswith momentum.
\newblock In \emph{IEEE Conference on Computer Vision and Pattern Recognition
  (CVPR)}, 2018.

\bibitem[Goodfellow et~al.(2015)Goodfellow, Shlens, , and
  Szegedy]{Goodfellow2015}
Goodfellow, I.~J., Shlens, J., , and Szegedy, C.
\newblock Explaining and harnessing adversarial examples.
\newblock In \emph{International Conference on Learning Representations
  (ICLR)}, 2015.

\bibitem[Ilyas et~al.(2019)Ilyas, Santurkar, Tsipras, Engstrom, Tran, and
  Madry]{Ilyas2019}
Ilyas, A., Santurkar, S., Tsipras, D., Engstrom, L., Tran, B., and Madry, A.
\newblock Adversarial examples are not bugs, they are features.
\newblock In \emph{Advances in Neural Information Processing Systems
  (NeurIPS)}, 2019.

\bibitem[Kang et~al.(2020)Kang, Sun, Hendrycks, Brown, and
  Steinhardt]{Kang2020}
Kang, D., Sun, Y., Hendrycks, D., Brown, T., and Steinhardt, J.
\newblock Testing robustness against unforeseen adversaries.
\newblock \emph{arXiv preprint arXiv:1908.08016v2}, 2020.

\bibitem[Kettunen et~al.(2019)Kettunen, H{\"a}rk{\"o}nen, and
  Lehtinen]{Kettunen2019}
Kettunen, M., H{\"a}rk{\"o}nen, E., and Lehtinen, J.
\newblock E-lpips: Robust perceptual image similarity via random transformation
  ensembles.
\newblock \emph{arXiv preprint arXiv:1906.03973}, 2019.

\bibitem[Kim et~al.(2020)Kim, Tack, and Hwang]{Kim2020}
Kim, M., Tack, J., and Hwang, S.~J.
\newblock Adversarial self-supervised contrastive learning.
\newblock In \emph{Advances in Neural Information Processing Systems
  (NeurIPS)}, 2020.

\bibitem[Krizhevsky \& Hinton(2009)Krizhevsky and Hinton]{Krizhevsky2009}
Krizhevsky, A. and Hinton, G.
\newblock Learning multiple layers of features from tiny images.
\newblock Technical report, 2009.

\bibitem[Kurakin et~al.(2017)Kurakin, Goodfellow, and Bengio]{Kurakin2017}
Kurakin, A., Goodfellow, I.~J., and Bengio, S.
\newblock Adversarial examples in the physical world.
\newblock In \emph{ICLR Workshop}, 2017.

\bibitem[Laidlaw \& Feizi(2019)Laidlaw and Feizi]{Laidlaw2019}
Laidlaw, C. and Feizi, S.
\newblock Functional adversarial attacks.
\newblock In \emph{Advances in Conference on Neural Information Processing
  Systems (NeurIPS)}, 2019.

\bibitem[Laidlaw et~al.(2020)Laidlaw, Singla, and Feizi]{Laidlaw2020}
Laidlaw, C., Singla, S., and Feizi, S.
\newblock Perceptual adversarial robustness: Defense against unseen threat
  models.
\newblock \emph{arXiv preprint arXiv:2006.12655}, 2020.

\bibitem[Lehman(2019)]{SegmentCifar}
Lehman, C.
\newblock Learning to segment {CIFAR10}, 2019.
\newblock URL
  \url{https://charlielehman.github.io/post/weak-segmentation-cifar10/}.

\bibitem[Lin et~al.(2018)Lin, Zhang, Hsu, Skach, Haque, Tang, and
  Mars]{Lin2018}
Lin, S.~C., Zhang, Y., Hsu, C.~H., Skach, M., Haque, M.~E., Tang, L., and Mars,
  J.
\newblock The architectural implications of autonomous driving: Constraints and
  acceleration.
\newblock In \emph{International Conference on Architectural Support for
  Programming Languages and Operating Systems (ASPLOS)}, 2018.

\bibitem[Madry et~al.(2018)Madry, Makelov, Schmidt, Tsipras, and
  Vladu]{Madry2018}
Madry, A., Makelov, A., Schmidt, L., Tsipras, D., and Vladu, A.
\newblock Towards deep learning models resistant to adversarial attacks.
\newblock In \emph{International Conference on Learning Representations
  (ICLR)}, 2018.

\bibitem[Maini et~al.(2020)Maini, Wong, and Kolter]{Miani2020}
Maini, P., Wong, E., and Kolter, J.~Z.
\newblock Adversarial robustness against the union of multiple perturbation
  models.
\newblock \emph{arXiv preprint arXiv:1909.04068v2}, 2020.

\bibitem[Milgrom \& Segal(2002)Milgrom and Segal]{Milgrom2002}
Milgrom, P. and Segal, I.
\newblock Envelope theorems for arbitrary choice sets.
\newblock \emph{Econometrica}, 70\penalty0 (2):\penalty0 583--601, 2002.

\bibitem[Modas et~al.(2019)Modas, Moosavi-Dezfooli, and
  Frossard]{Apostolos2019}
Modas, A., Moosavi-Dezfooli, S., and Frossard, P.
\newblock Sparsefool: a few pixels make a big difference.
\newblock In \emph{IEEE Conference on Computer Vision and Pattern Recognition
  (CVPR)}, 2019.

\bibitem[Moosavi-Dezfooli et~al.(2016)Moosavi-Dezfooli, Fawzi, and
  Frossard]{Moosavi2016}
Moosavi-Dezfooli, S., Fawzi, A., and Frossard, P.
\newblock Deepfool: a simple and accurate method to fool deep neural networks.
\newblock In \emph{IEEE Conference on Computer Vision and Pattern Recognition
  (CVPR)}, 2016.

\bibitem[Rony et~al.(2019)Rony, Hafemann, Oliveira, Ayed, Sabourin, and
  Granger]{Rony2019}
Rony, J., Hafemann, L.~G., Oliveira, L.~S., Ayed, I.~B., Sabourin, R., and
  Granger, E.
\newblock Decoupling direction and norm for efficient gradient-based {L2}
  adversarial attacks and defenses.
\newblock In \emph{Advances in Neural Information Processing Systems
  (NeurIPS)}, 2019.

\bibitem[Rony et~al.(2020)Rony, Granger, Pedersoli, and Ayed]{Rony2020}
Rony, J., Granger, E., Pedersoli, M., and Ayed, I.~B.
\newblock Augmented lagrangian adversarial attacks.
\newblock \emph{arXiv preprint arXiv:2011.11857}, 2020.

\bibitem[Russakovsky et~al.(2015)Russakovsky, Deng, Su, Krause, Satheesh, Ma,
  Huang, Karpathy, Khosla, Bernstein, Berg, and Fei-Fei]{ILSVRC152015}
Russakovsky, O., Deng, J., Su, H., Krause, J., Satheesh, S., Ma, S., Huang, Z.,
  Karpathy, A., Khosla, A., Bernstein, M., Berg, A.~C., and Fei-Fei, L.
\newblock {ImageNet Large Scale Visual Recognition Challenge}.
\newblock \emph{International Journal of Computer Vision (IJCV)}, 115\penalty0
  (3):\penalty0 211--252, 2015.
\newblock \doi{10.1007/s11263-015-0816-y}.

\bibitem[Szegedy et~al.(2014)Szegedy, Zaremba, Sutskever, Bruna, Erhan,
  Goodfellow, and Fergus]{Szegedy2014}
Szegedy, C., Zaremba, W., Sutskever, I., Bruna, J., Erhan, D., Goodfellow, I.,
  and Fergus, R.
\newblock Intriguing properties of neural networks.
\newblock In \emph{International Conference on Learning Representations
  (ICLR)}, 2014.

\bibitem[Tramèr \& Boneh(2019)Tramèr and Boneh]{Trame2019}
Tramèr, F. and Boneh, D.
\newblock Adversarial training and robustness for multiple perturbations.
\newblock In \emph{Advances in Conference on Neural Information Processing
  Systems (NeurIPS)}, 2019.

\bibitem[Tsipras et~al.(2019)Tsipras, Santurkar, Engstrom, Turner, and
  Madry]{tsipras2019}
Tsipras, D., Santurkar, S., Engstrom, L., Turner, A., and Madry, A.
\newblock Robustness may be at odds with accuracy.
\newblock In \emph{International Conference on Learning Representations
  (ICLR)}, 2019.

\bibitem[Uesato et~al.(2018)Uesato, O'Donoghue, Kohli, and van~den
  Oord]{Uesato2018}
Uesato, J., O'Donoghue, B., Kohli, P., and van~den Oord, A.
\newblock Adversarial risk and the dangers of evaluating against weak attacks.
\newblock In \emph{International Conference on Machine Learning (ICML)}, 2018.

\bibitem[Wang \& Deng(2020)Wang and Deng]{Wang2020}
Wang, M. and Deng, W.
\newblock Deep face recognition: A survey.
\newblock \emph{arXiv preprint arXiv:1804.06655v9}, 2020.

\bibitem[Wang et~al.(2020)Wang, Zou, Yi, Bailey, Ma, and Gu]{Wang2020Improving}
Wang, Y., Zou, D., Yi, J., Bailey, J., Ma, X., and Gu, Q.
\newblock Improving adversarial robustness requires revisiting misclassified
  examples.
\newblock In \emph{International Conference on Learning Representations
  (ICLR)}, 2020.

\bibitem[Wong et~al.(2020)Wong, Schmidt, and Kolter]{Wong2020}
Wong, E., Schmidt, F.~R., and Kolter, J.~Z.
\newblock Wasserstein adversarial examples via projected sinkhorn iterations.
\newblock \emph{arXiv preprint arXiv:1902.07906v2}, 2020.

\bibitem[Xiao et~al.(2018)Xiao, Zhu, Li, He, Liu, and Song]{xiao2018}
Xiao, C., Zhu, J.-Y., Li, B., He, W., Liu, M., and Song, D.
\newblock Spatially transformed adversarial examples.
\newblock In \emph{International Conference on Learning Representations
  (ICLR)}, 2018.

\bibitem[Zagoruyko \& Komodakis(2017)Zagoruyko and Komodakis]{Zagoruyko2017AT}
Zagoruyko, S. and Komodakis, N.
\newblock Paying more attention to attention: Improving the performance of
  convolutional neural networks via attention transfer.
\newblock In \emph{International Conference on Learning Representations
  (ICLR)}, 2017.

\bibitem[Zhang et~al.(2019)Zhang, Yu, Jiao, Xing, Ghaoui, and
  Jordan]{Zhang2019}
Zhang, H., Yu, Y., Jiao, J., Xing, E.~P., Ghaoui, L.~E., and Jordan, M.~I.
\newblock Theoretically principled trade-off between robustness and accuracy.
\newblock In \emph{International Conference on Machine Learning (ICML)}, 2019.

\bibitem[Zhang et~al.(2018)Zhang, Isola, Efros, Shechtman, and Wang]{zhang2018}
Zhang, R., Isola, P., Efros, A.~A., Shechtman, E., and Wang, O.
\newblock The unreasonable effectiveness of deep features as a perceptual
  metric.
\newblock In \emph{The IEEE Conference on Computer Vision and Pattern
  Recognition (CVPR)}, 2018.

\bibitem[{Zhou Wang} et~al.(2004){Zhou Wang}, {Bovik}, {Sheikh}, and
  {Simoncelli}]{Wang2004}
{Zhou Wang}, {Bovik}, A.~C., {Sheikh}, H.~R., and {Simoncelli}, E.~P.
\newblock Image quality assessment: from error visibility to structural
  similarity.
\newblock \emph{IEEE Transactions on Image Processing}, 2004.

\end{thebibliography}
\bibliographystyle{icml2021}
%\iffalse
\newpage
\null\newpage
\appendix
\section{Flowers recognition}
As demonstrated earlier, our method performs better than the others against unseen attacks on CIFAR-10 and ImageNet-100 datasets. Nevertheless, a big concern about all of the methods is that they overfit to these two datasets. In other words, these two datasets have mainly been used for  the evaluations, and the satisfactory performance of a method may be limited to these datasets. To address this concern, we examine our method against PGD attacks on another dataset. We set a high bar, by comparing our method against PGD-$\ell_\infty$ and PGD-$\ell_2$ trained model on $\ell_\infty$ and $\ell_2$ balls, respectively. 

\begin{figure}[h]
    \centering
    \includegraphics[scale=0.5]{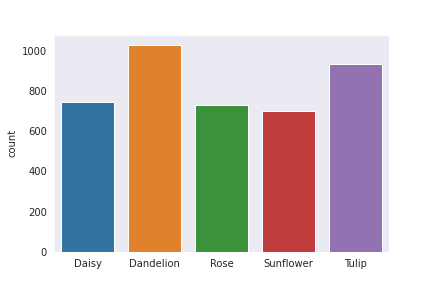}
    \vskip -0.1in
    \caption{Distribution of various classes in the Flower dataset.}
    \label{fig:Count_flower}
\end{figure}

\begin{figure}[h]
    \centering
    \includegraphics[scale=0.58]{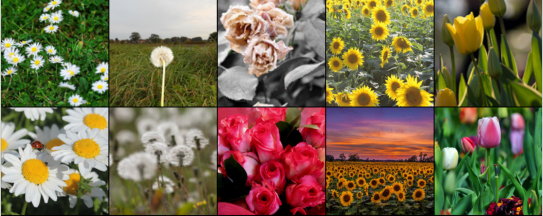}
    \vskip -0.1in
    \caption{Sample images from the Flower dataset. The columns are Daisy, Dandelion, Rose, Sunflower, and Tulip, respectively.}
    \label{fig:flower}
\end{figure}

It is more beneficial to select a dataset with different types of images than the mentioned datasets. It is also preferable that the data be almost balanced and its images have an appropriate resolution. For this purpose, the Flower dataset\footnote{\url{https://www.kaggle.com/alxmamaev/flowers-recognition}} is used that contains five different types of flowers (Daisy, Dandelion, Rose, Sunflower, and Tulip). The resolution of the images is about 320x240. This dataset includes some irrelevant images, which we removed for a more accurate evaluation. Looking at the number of samples available in each class in Fig. \ref{fig:Count_flower}, we realize that the dataset is sufficiently balanced. Some samples in this dataset are shown in Fig. \ref{fig:flower}.

\begin{figure}[h]
    \centering
    \includegraphics[scale=0.6]{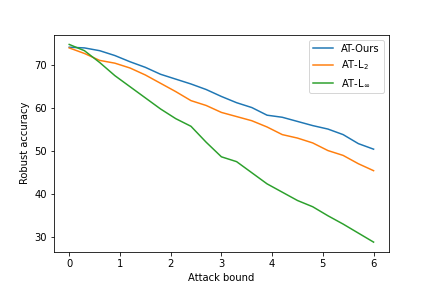}
    \vskip -0.1in
    \caption{Evaluating adversarially trained models on the Flower dataset against PGD-$\ell_2$ with various bounds. Allowable $\ell_2$ norm of the perturbation is gradually increased from zero to six.}
    \label{fig:flower_l2}
\end{figure}

\begin{figure}[h!]
    \centering
    \includegraphics[scale=0.6]{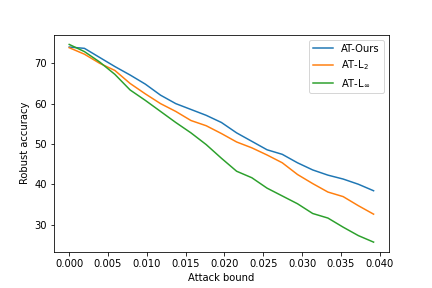}
    \vskip -0.1in
    \caption{Evaluating adversarially trained models on the Flower dataset against PGD-$\ell_\infty$ with various bounds. Allowable $\ell_\infty$ norm of the perturbation is gradually increased from zero to $10/255$.}
    \label{fig:flower_linf}
\end{figure}

We trained a ResNet-18 model to classify the flowers using the same training setting as in ImageNet-100. We evaluated it against the PGD attacks with various bounds. Results are shown in Fig. \ref{fig:flower_l2} and Fig. \ref{fig:flower_linf}, which demonstrate the better performance of our method than training with the same attacks being used at the test time. Note the outperforming such baselines is more challenging than comparing with other methods against the unseen attacks.

\section{Adversarial evaluations}
As far as possible, attempts were made to perform the evaluations as comprehensively as possible. Various attacks are used to obtain the robust accuracy of different methods against unforeseen attacks and attacks used during training to make the assessments entirely fair. Since our context evaluated methods against unseen attacks, we only used PGD-$\ell_\infty/\ell_2$ and didn't use stronger versions of PGD. Although if we had used stronger attacks such as AutoAttack \cite{croce2020}, the results would not have changed. To illustrate this point, we checked the accuracy of our method and PAT's method as the method that has the closest clean and unseen accuracy to ours, against standard AutoAttack in CIFAR-10 and ImageNet-100 dataset. The results are reported in Table \ref{tab:AA} that demonstrates a similar trend in AutoAttack accuracies to PGD attacks.

Also, to show one aspect of the variety of selected attacks, we examined the average $\ell_2$ norm of selected attacks on our CIFAR-10 model, which is reported in Table \ref{tab:L2_norm} and shows the chosen attacks' $\ell_2$ norm diversity. Furthermore, the slightly larger $\ell_2$ norm of the attacks than our training attack is another reason for the selected attacks' fairness. However, it confirms that our contribution is also about removing the LPIPS bias.

\begin{table}[htbp]
  \centering
  \begin{small}
  \caption{Comparison of AutoAttack (AA) with PGD (perturbation budgets are similar to Tables \ref{CIFAR10 results} and \ref{ImageNet results}). Both show that our model is more robust. }
  \vskip 0.1in
  \renewcommand{\arraystretch}{1.3} % Default value: 1
    \begin{tabular}{c|c|cc|cc}
    \multirow{2}[3]{*}{Dataset} & \multirow{2}[3]{*}{Model} & \multicolumn{2}{c|}{$\ell_\infty$} & \multicolumn{2}{c}{$\ell_2$} \\
\cmidrule{3-6}          &       & AA & PGD   & AA & PGD \\
    \midrule
    \multirow{2}[2]{*}{CIFAR-10} & Ours  & 38.47 & 48.12 & 44.98 & 54.17 \\
          & PAT   & 33.75 & 43.19 & 39.21 & 46.19 \\
    \midrule
    \multirow{2}[1]{*}{ImageNet-100} & Ours  &   40.23    & 46.47 &   39.80    & 46.59 \\
          & PAT   &    37.18   & 43.14 &    37.51   & 43.29 \\
    \end{tabular}%
  \label{tab:AA}%
  \end{small}
\end{table}%

\begin{table}[h]
  \centering
  \begin{small}
  \caption{Average $\ell_2$ norm of attacks on our CIFAR-10 model.}
    \vskip 0.1in
    \renewcommand{\arraystretch}{1.3} % Default value: 1
    \begin{tabular}{cc|cc}
    Ours  & 1.31  & Elastic & 7.87 \\
    \midrule
    LPA   & 2.45  & Jpeg  & 2.07 \\
    \midrule
    $\ell_\infty$  & 1.56  & Noise & 12.39 \\
    \midrule
    $\ell_2$ & 1.00     & Blur  & 5.68 \\
    \midrule
    Recolor & 2.18  & PGD$_0$  & 7.87 \\
    \midrule
    StAdv & 3.67  &       &  \\
    \end{tabular}%
  \label{tab:L2_norm}%
  \end{small}
\end{table}%
%\fi
\end{document}